\documentclass[sigconf]{acmart}
\usepackage{enumitem}
\usepackage{graphicx}
\usepackage{pifont}
\usepackage{xcolor}
\usepackage{soul}
\usepackage{subcaption}
\usepackage{subfiles}
\newcommand{\hlc}[2][yellow]{{%
    \colorlet{foo}{#1}%
    \sethlcolor{foo}\hl{#2}}%
}

\AtBeginDocument{%
  }

\setcopyright{acmlicensed}
\copyrightyear{2018}
\acmYear{2018}
\acmDOI{XXXXXXX.XXXXXXX}
\acmISBN{978-1-4503-XXXX-X/18/06}

\begin{document}

\title{Towards Zero-Shot Annotation of the Built Environment with Vision-Language Models (Vision Paper)}

\author{Bin Han}
\email{bh193@uw.edu}
\affiliation{%
  \institution{University of Washington}
  \city{Seattle}
  \country{USA}}
  
\author{Yiwei Yang}
\email{yanyiwei@uw.edu}
\affiliation{%
  \institution{University of Washington}
  \city{Seattle}
  \country{USA}}
 
\author{Anat Caspi}
\email{caspian@cs.washington.edu}
\affiliation{%
  \institution{University of Washington}
  \city{Seattle}
  \country{USA}}
  
\author{Bill Howe}
\email{billhowe@uw.edu}
\affiliation{%
  \institution{University of Washington}
  \city{Seattle}
  \country{USA}}

\begin{abstract}
    Equitable urban transportation applications require high-fidelity digital representations of the built environment: not just streets and sidewalks, but bike lanes, marked and unmarked crossings, curb ramps and cuts, obstructions, traffic signals, signage, street markings, potholes, and more.  
    Direct inspections and manual annotations are prohibitively expensive at scale. Conventional machine learning methods require substantial annotated training data for adequate performance. In this paper, we consider vision language models as a mechanism for annotating diverse urban features from satellite images, reducing the dependence on human annotation to produce large training sets.  While these models have achieved impressive results in describing common objects in images captured from a human perspective, their training sets are less likely to include strong signals for esoteric features in the built environment, and their performance in these settings is therefore unclear. 
    We demonstrate proof-of-concept combining a state-of-the-art vision language model and variants of a prompting strategy that asks the model to consider segmented elements independently of the original image. 
    Experiments on two urban features --- stop lines and raised tables --- show that while direct zero-shot prompting correctly annotates nearly zero images, the pre-segmentation strategies can annotate images with near 40\% intersection-over-union accuracy.  
    We describe how these results inform a new research agenda in automatic annotation of the built environment to improve equity, accessibility, and safety at broad scale and in diverse environments. 
\end{abstract}

\ccsdesc[500]{Computing methodologies~Scene understanding}
\ccsdesc[500]{Computing methodologies~Visual content-based indexing and retrieval}
\ccsdesc[500]{Computing methodologies~Visual inspection}
\ccsdesc[500]{Computing methodologies~Image segmentation}
\ccsdesc[500]{Computing methodologies~Interest point and salient region detections}

\keywords{Large Language Model, Multi-Modality Model, Image Segmentation,  Urban Computing, Urban Data Annotation}

\maketitle

\vspace{-0.3cm}
\section{Introduction}\label{sec:intro}
\begin{figure*}[!ht]
    \centering
    \includegraphics[width=\linewidth]{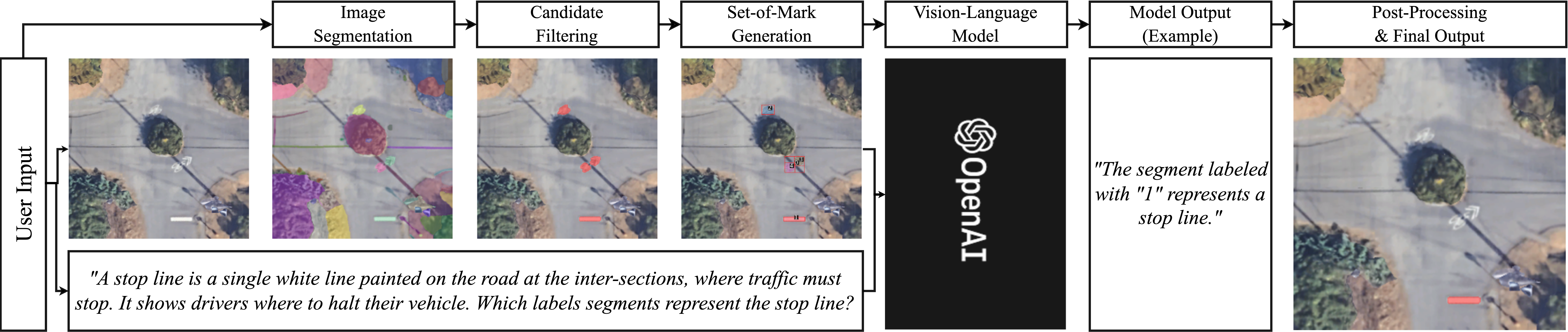}
    \vspace{-0.75cm}
    \caption{Pipeline of our proposed automated annotation process. Users input a pair of (satellite image, annotation guidance). The image will go through a set of processes including segmentation, filtering, and set-of-mark generation. Then the image and guidance will go through a vision-language model, the output of which is post-processed to produce the final annotation results. The procedure requires no fine-tuning, and can be applied on different features with minimal adjust on the guidance.}
    \label{fig:pipeline}
    \vspace{-0.35cm}
\end{figure*}

The inaccessible urban infrastructure reinforces systemic exclusion of people with disabilities and negatively impacts public health and overall quality of life for everyone \cite{public_health}.  As laws evolve to make the urban environment more accessible, the digital representation must also evolve to include not just roads and sidewalks, but the hundreds of features that enable safe and equitable mobility: curb ramps, walk signals, tactile paving, crosswalks, light poles, benches, and many more~\cite{mendoza2023}.  For example, mobility applications that provide accessible pedestrian routes rely on accurate, comprehensive annotations of the dynamic urban infrastructure~\cite{bolten19}.

The gold standard for data collection involves in-person inspections, typically administered by local governments with the help of community volunteers or contractors \cite{froehlich2022future}. Human annotations are high quality but are labor-intensive, costly, and difficult to scale. For instance, in 2017, the city of Seattle employed 14 people for their first-ever audit of 2,300 miles of sidewalks, costing \$400,000 and identifying 92,000 uplifts, 38,000 surface problems, and 20,000 obstructions\cite{froehlich2022future}. Sensing platforms deployed to navigate the built environment can partially automate annotation via machine learning, but these initiatives are also expensive to scale~\cite{robot}.
Aerial imagery offers nearly complete coverage at low cost, as long as the computer vision models to extract the relevant features can be trained. In practice, model development tends to focus on the needs of the "typical" traveler, e.g., roads~\cite{lidar} and in a few cases sidewalks~\cite{hosseini2023mapping,zhang2023ape}.  Expanding coverage to the long tail of accessibility features is rarely viable due to the cost in obtaining training data.

Pre-trained vision-language models (VLMs) (e.g., \cite{clip, llava}), exhibit impressive multi-modal scene understanding that can be used, among many other tasks, to identify objects in images~\cite{redcircle}. If these models could be used
to bootstrap feature annotation in the built environment, a more complete digital record could be developed enabling equity and accessibility at scale~\cite{howe2022integrative}.  However, these models are trained on text paired with images from the internet, such that common objects in everyday (US-centric) experiences tend to be dominant. For atypical scenes and images, including satellite images, performance remains to be tested. 

In this vision paper, we consider using VLMs with aerial imagery as a source of low-cost, extensible annotations for arbitrary features in the built environment.  By reducing the cost of producing annotations to near zero, mobility applications can keep pace with an evolving building code and commensurate new construction, faithfully representing the complete and current built environment.  With this capability, we envision a larger roadmap to scale this annotation process to any feature in any region, dramatically lowering costs for data acquisition and therefore enable a broad class of community-focused, AI-based mobility applications that have previously been too specialized to justify the cost. The ability to make services for specialized needs as economically viable as services for typical needs is the foundation of equity.

To demonstrate the proof-of-concept, we developed a procedure borrowing techniques from visual grounding~\cite{vg1, vg2}, visual prompting~\cite{vp1}, and set-of-mark prompting~\cite{som} that generates bounding box annotations on satellite images based only on high-level descriptions of features. The procedure requires no fine-tuning or even labeled examples, demonstrating that closed, general purpose models are potentially viable even in a zero-shot setting. 

We use the a state-of-the-art VLM (GPT-4o\footnote{\url{https://platform.openai.com/docs/models/gpt-4o}}) and a general-purpose segmentation model (SAM \cite{sam}). The SAM model segments a satellite image into a gallery of candidate objects, from which the GPT-4o model can select the appropriate feature of interest (see Figure \ref{fig:pipeline} for details). We test our procedure on two urban features: stop lines (white lines indicating where cars should stop at an intersection, which are important for safely deriving pedestrian routes through unmarked crossings) and raised tables (speed control features that can be used to estimate traffic flow and therefore pedestrian risk.) We consider multiple variants of this procedure, showing that presenting the segmented "options" to the model in different ways can improve performance. 
Quantitatively, we show that a simple direct prompting strategy fails dramatically at this task, while the proposed procedures achieve on average about 40\% intersection-over-union accuracy. The takeaway is that this problem is difficult, but not impossible.
Qualitatively, we visualize several correct and incorrect annotations to inform a discussion of the challenges and vision. Overall, our contributions are:
\begin{itemize}[leftmargin=0.12in]
    \item We describe a prompting procedure for generating annotations urban features from aerial imagery using a VLM, adapting segmentation based annotations methods to this context. 
    \item We evaluate the procedure on two urban features, finding that while simple methods fail, the proposed methods are promising and viable, which suggests an important avenue for research. 
    \item Given these results, we  discuss the current challenges faced by the automated process and describe how a complete, accurate, and publicly available data record may now be feasible and consider next steps for the research community to realize this vision.
\end{itemize}
\vspace{-0.25cm}

\section{Related Work}\label{sec:related_word}
\begin{table*}[!ht]
    \footnotesize
    \centering
    \tabcolsep=0.35cm
    \renewcommand*{\arraystretch}{0.85}
    \begin{tabular}{|c|p{7cm}|p{7cm}|} 
        \hline
        \textbf{Object} & \textbf{DP} & \textbf{SoM-(NC,IC,Comb)} \\ \hline
        Stop Line & \hlc[pink]{A stop line is a single white line painted on the road at intersections where traffic must stop. It shows drivers where to halt their vehicles.} \hlc[cyan!50]{Please identify the bounding box of the stop line in the image in the format of (xtl, ytl, xbr, ybr).} & \hlc[pink]{A stop line is a single white line painted on the road at intersections where traffic must stop. It shows drivers where to halt their vehicles.}  \hlc[cyan!50]{Which labeled images represent the stop line?}\\ \hline
        
        Raised Table & \hlc[pink]{A raised table usually covers the entire width of the crosswalk. It is typically painted with triangular arrows in white color.} \hlc[cyan!50]{Please identify the bounding box of the stop line in the image in the format of (xtl, ytl, xbr, ybr).} & \hlc[pink]{A raised table usually covers the entire width of the crosswalk. It is typically painted with triangular arrows in white color.} \hlc[cyan!50]{Which labeled images represent the raised table?} \\ \hline
    \end{tabular}
    \caption{Text prompts for the VLM model to annotate different features of interest. \hlc[pink]{Pink} highlighted texts represent the descriptive information of the features. \hlc[cyan!50]{Blue} highlighted texts represent the general task questions.} 
    \label{tab:prompts}
    \vspace{-0.85cm}
\end{table*}

\textbf{Urban Data Collection:} 
\cite{zhang2023ape} introduced a novel dataset comprising aerial satellite imagery, street maps, and rasterized annotations of sidewalks, crossings, and corner bulbs in cities. Even after years of work, existing models recognize only the three most important features of the pedestrian network. These models require expansive, diverse training data to perform well; our focus is on bootstrapping a process for collecting this training data for rare cases. 

\noindent\textbf{Visual Grounding \& Visual Prompting:} Our proposed system is highly relevant to visual grounding task and visual prompting technique. Visual grounding task in computer vision aims to locate the most relevant object or region in an image based on the specific task description in natural language \cite{vg1, vg2}. Visual prompting is a technique that uses visual cues or examples to guide multi-modal AI models in understanding and responding to visual tasks, similar to how textual prompts are used to instruct LLMs \cite{vp1}. Our proposed procedure is based on Set-of-Mark (SoM) prompting \cite{som}, leveraging off-the-shelf segmentation models to partition images into smaller regions, and overlays each region with a mark to provide visual hints for the model. Their work demonstrated extraordinary visual grounding ability of GPT-4.
\vspace{-0.25cm}

\section{Prompting Procedure}\label{sec:method}

\begin{figure}[h]
    \centering
    \vspace{-0.5cm}
    \includegraphics[width=0.95\linewidth]{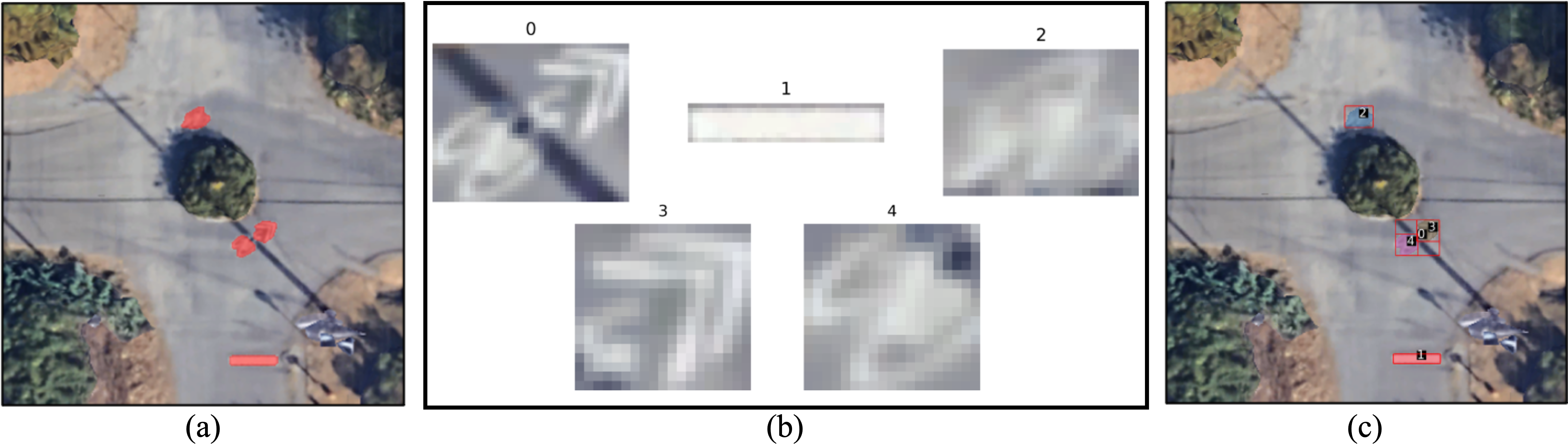}
    \vspace{-0.35cm}
    \caption{SoM generation scenarios: (a) Filtered candidates (b) No-Context: Candidate objects are presented separately in a new image. (c) In-Context: Candidate objects are labeled with numbers and bounding boxes within the original image.}
    \label{fig:som}
    \vspace{-0.5cm}
\end{figure}

Our proposed procedure is shown in Figure \ref{fig:pipeline}. It works as follows:
\begin{itemize}[leftmargin=0.12in]
    \item \textbf{User Input:} a pair of (satellite image, annotation guidance).
    \item \textbf{Image Segmentation:} The image is segmented into a set of objects using a general-purpose model, ignoring the prompt.     
    \item \textbf{Candidate Filtering:} We apply heuristic filters to eliminate irrelevant components and narrow down the candidate space. In our experiments, we employ color and area size filters, excluding green, yellow, brown, and red elements, as well as components smaller than specific sizes\footnote{200 pixel area for stop lines and 400 pixel area for raised tables.}. In practice, different parameter values and additional filter criteria can be applied based on the specific urban features and expert knowledge.    
    \item \textbf{Set-of-Mark (SoM) Generation:} After filtering, we generate an identifying \textit{mark} for each component~\cite{som} to enhance the vision-language model's recognition of each object. We test three different strategies as shown in Figure \ref{fig:som}:
    \begin{itemize}[leftmargin=0.12in]
        \item \textbf{No-Context (NC)}: the candidates are extracted and presented as distinct, unrelated objects embedded in a plain white background, with each component labeled with a number.  
        \item \textbf{In-Context (IC)}: we label the candidates with numbers and bounding boxes within the original image.   
        \item \textbf{Combination (Comb)}: We generate both no- and in-context images and provide both to the VLM.
    \end{itemize}
    \item \textbf{VLM \& Post-Processing:} the image from the SoM generation process is sent to the VLM with the textual annotation guidance. The text prompt consists of 1) a description of the feature of interest (e.g., stop line, raised table) and what specific features it possesses and 2)
    the instructions for annotating the image (see Table \ref{tab:prompts}).
    The output of the model is post-processed: the model responds with the mark indicating the selected candidate and we return the corresponding bounding box. 
\end{itemize}

\section{Evaluation \& Results}

\begin{figure*}[t!]
    \centering
    \begin{subfigure}[t]{0.5\textwidth}
        \centering
        \includegraphics[width=\linewidth]{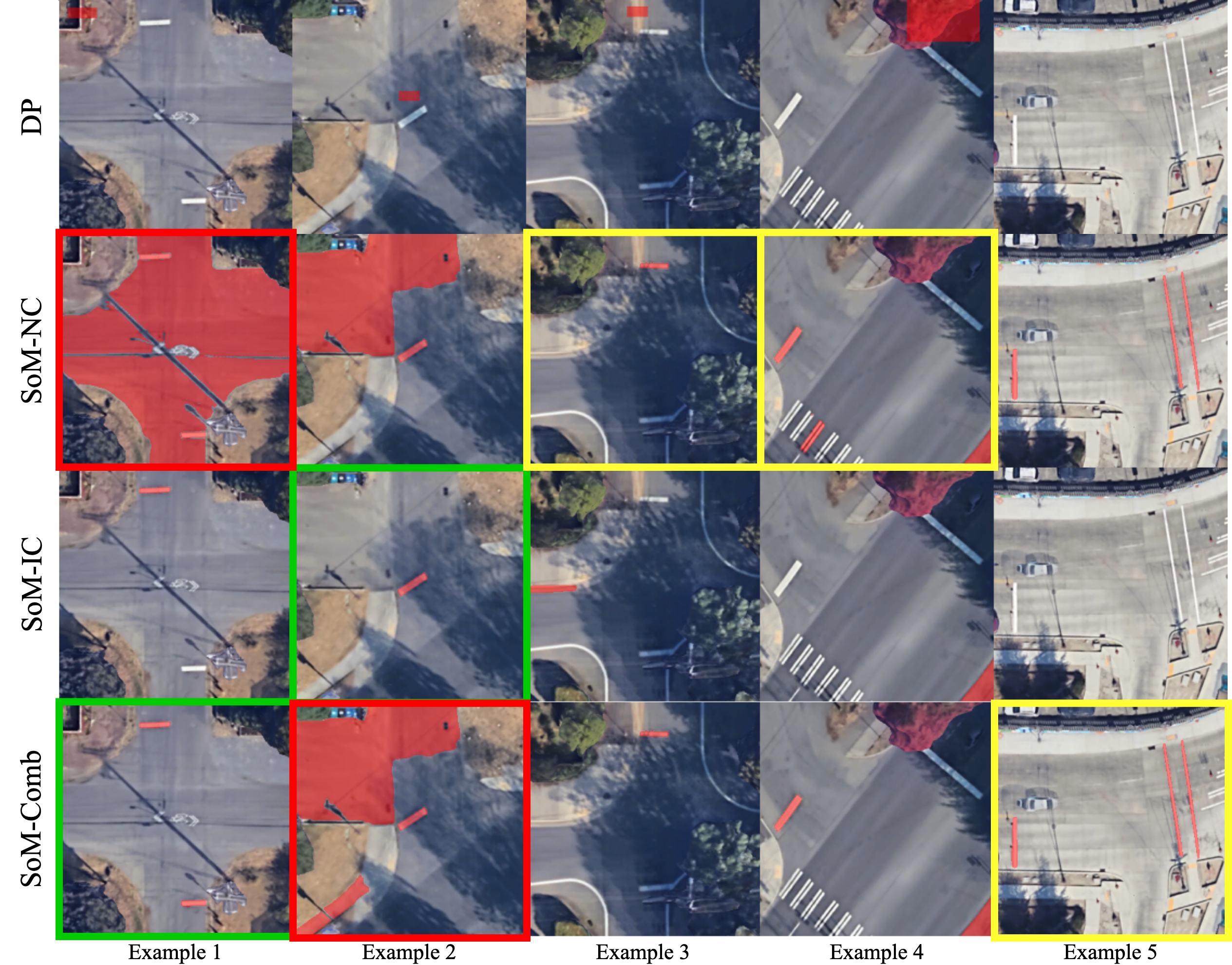}
    \end{subfigure}%
    \begin{subfigure}[t]{0.5\textwidth}
        \includegraphics[width=\linewidth]{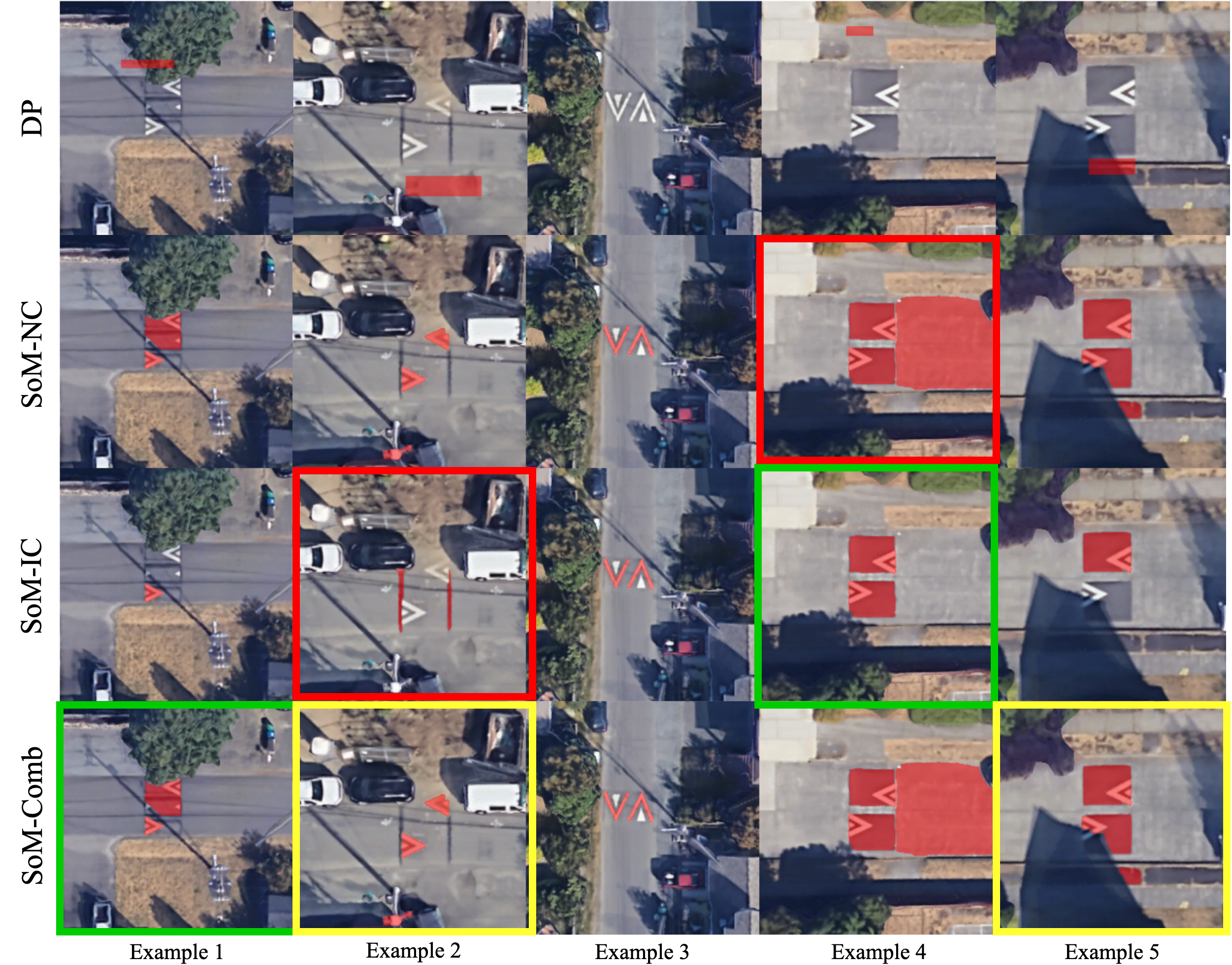}
    \end{subfigure}
    \vspace{-0.75cm}
    \caption{Left -- Examples of annotated stop lines. Right -- Examples of annotated raised tables. Red regions in each image are the segmented objects. \textcolor{green}{Green} and \textcolor{yellow}{yellow} outlines indicate perfect and approximate annotations, respectively. A \textcolor{red}{Red} outline indicate inaccurate annotations.}
    \label{fig:examples}
    \vspace{-0.5cm}
\end{figure*}

To evaluate this procedure, we manually cropped and annotated eight satellite images that contains stop lines and ten images that contain raised tables from Google Earth. As a baseline, we compare against a \textbf{Direct Prompting (DP)} method: given an  image, we directly ask GPT-4o (without SAM) to annotate the feature of interest and return the bounding boxes. 

Quantitatively, we evaluate the annotations using Intersection-over-Union (IoU) metric: $IoU = \frac{G\cap P}{G \cup P}$, where $G$ is the ground truth annotation and $P$ is the annotation from our procedure. The results are presented in Table \ref{tab:quantitative}. We observe that 1) Direct prompting is completely  ineffective, achieving essentially zero overlap with ground truth.  The annotations reflect very little understanding of either the instructions or the image, which is consistent with prior work~\cite{som}.2) Adding visual marks significantly improves performance, achieving 24\%-40\% IoU depending on the variant.  In-context improves over no-context by 47.3\% and 26.4\% on two features respectively, and providing both forms yields an additional slight improvement.

Qualitatively, we visualize a few annotated examples to demonstrate the feasibility of the automated process. We also mark some flawed examples to be discussed in the next section \S\ref{sec:challenges}. From the visualizations, we observe 1) the failure of direct prompting, indicating the problem is difficult and not amenable to trivial solutions; 2) given visual marks, the VLM model can always at least partially identify (highlighted with yellow boxes), and occasionally fully identify the features (highlighted with green boxes), suggesting the problem is feasible for other types of features; and 3) among the three annotation scenarios, there is no consistent winner, motivating further research. 
\begin{table}[ht]
    \vspace{-0.25cm}
    \centering
    \tabcolsep=0.22cm
    \renewcommand*{\arraystretch}{1}
    \begin{tabular}{|c|c|c|c|c|} 
        \hline
        \textbf{Feature} & \textbf{DP} & \textbf{SoM-NC} & \textbf{SoM-IC} & \textbf{SoM-Comb} \\ \hline
        Stop Line & 0.0000 & 0.2483 & 0.3354 & 0.3657 \\ 
        Raised Table & 0.0190 & 0.3315 & 0.4069 & 0.4189 \\
        \hline
    \end{tabular}
    \caption{Quantitative evaluation using IoU(\rotatebox{90}{\color{green!80!black}\ding{225}}) metric. } 
    \label{tab:quantitative}
    \vspace{-0.85cm}
\end{table}
\vspace{-0.25cm}

\section{Challenges \& Future Direction}\label{sec:challenges}
Inspecting or manually annotating street objects is labor-intensive and time-consuming. Our proposed procedure reduces the annotation process to a) describing a typical representation of the feature in question, and potentially 2) providing simple heuristics for filtering candidates based on color, size, other basic image properties.  We evaluate in the zero-shot setting to reduce the dependency on finding representative examples --- a user can directly apply domain knowledge to generate training data from raw satellite images.  
This procedure can produce many thousands of annotated, not all of which will be correct.  In ongoing work, we are using these bootstrapped images to train more specialized vision models; we anticipate that the downstream model will tend to be robust to incorrect annotations, as long as most are roughly correct.   This procedure can be completed in a few minutes, allowing users to rapidly produce annotated images for any  feature they can describe.   

Our experiments suggest a solution is feasible but expose several challenges and motivate directions for further work:

\begin{itemize}[leftmargin=0.12in]
    \item \textbf{Noisy/Missing Segmentation}. The annotation results heavily depend on the quality of the segmented components. 1) Since SAM is a general-purposed segmentation model, it often produces irrelevant segments. Currently, we prune the space of candidates heuristically to improve performance, but these heuristics may need to be developed for each feature individually, and it is unclear how to perform this task in a principled way.  
    2) SAM might miss part of or the entire feature. For example, part of the raised table is covered by the shadow, thus not being detected 
    To address both issues, we recommend studyingsegmentation models specialized for satellite images that can, e.g., ignore shadows.
    
    \item \textbf{Capability of Vision-Language Models}. The rapid improvements in VLMs mediate our results --- Anecdotally, the recently released GPT-4o dramatically outperformed the state of the art open VLM Llava2 and yet still makes mistakes suggesting limited comprehension of the built environment. For example, crossing lines oriented  parallel to the roadway are mistaken for stop lines (Figure \ref{fig:examples}, example 5, SoM-NC, yellow box) or misunderstand the task entirely (Figure \ref{fig:examples}, example 1, SoM-NC, red box). VLMs finetuned on satellite images may avoid these errors. 
    
    \item \textbf{High Variations}. Even with a zero temperature value (which makes outputs from the VLM more deterministic), GPT-4o can produce varying results for the same image and prompt inputs over time. This variability leads to inconsistent annotations, requiring multiple rounds to achieve reliable outcomes.
    
    \item \textbf{City-wise Generalizability}. Since the process relies on two general-purpose large AI models, it can potnetially generalize to any feature in any location. However, this generalization remains untested. Some features of interest will be significantly harder to recognize: potholes have irregular shapes and sizes; benches will be more often shaded or occluded.  
    Further, laws and standards vary across cities, changing the visible environment.
\end{itemize}
\vspace{-0.5cm}

\bibliographystyle{ACM-Reference-Format}
\bibliography{annotation}

\end{document}